\documentclass[runningheads]{llncs}
\usepackage{graphicx}
\usepackage{amssymb}
\usepackage{amsmath}
\usepackage{bm}
\usepackage{xcolor}
\usepackage{hyperref}
\usepackage{subcaption}
\usepackage[section]{placeins}

\begin{document}
\def\arraystretch{1.15}
\setlength\tabcolsep{.5em}
\title{Paying Per-label Attention for Multi-label Extraction from Radiology Reports}
\titlerunning{Paying Per-Label Attention for Multi-label Extraction}
\author{
Patrick Schrempf\inst{1,2}
\and
Hannah Watson\inst{1}
\and
Shadia Mikhael\inst{1}
\and
\\Maciej Pajak\inst{1}
\and
Mat\'{u}\v{s} Falis\inst{1}
\and
Aneta Lisowska\inst{1}
\and
Keith W. Muir\inst{3}
\and
\\David Harris-Birtill\inst{2}
\and
Alison Q. O'Neil\inst{1,4}
}
%
%
\authorrunning{P. Schrempf et al.}
\institute{Canon Medical Research Europe, Edinburgh, United Kingdom
\and
University of St Andrews, United Kingdom
\and
Institute of Neuroscience \& Psychology, University of Glasgow, United Kingdom
\and
University of Edinburgh, United Kingdom\\
\email{patrick.schrempf@eu.medical.canon}}
\maketitle              
\begin{abstract}
Training medical image analysis models requires large amounts of expertly annotated data which is time-consuming and expensive to obtain. Images are often accompanied by free-text radiology reports which are a rich source of information. In this paper, we tackle the automated extraction of structured labels from head CT reports for imaging of suspected stroke patients, using deep learning. Firstly, we propose a set of 31 labels which correspond to radiographic findings (e.g. hyperdensity) and clinical impressions (e.g. haemorrhage) related to neurological abnormalities. Secondly, inspired by previous work, we extend existing state-of-the-art neural network models with a label-dependent attention mechanism. Using this mechanism and simple synthetic data augmentation, we are able to robustly extract many labels with a single model, classified according to the radiologist's reporting (positive, uncertain, negative). This approach can be used in further research to effectively extract many labels from medical text.
\keywords{NLP \and Radiology report labelling \and BERT}
\end{abstract}
\section{Introduction}

Training medical imaging models requires large amounts of expertly annotated data which is time-consuming and expensive to obtain. Fortunately, medical images are often accompanied by free-text reports written by radiologists summarising their main \emph{findings} (what the radiologist sees in the image e.g. ``hyperdensity'') and \emph{impressions} (what the radiologist diagnoses based on the findings e.g. ``haemorrhage''). This information can be converted to structured labels which are used to train image analysis algorithms to detect the findings and to predict the impressions. Image-level labels have previously been provided to train image analysis algorithms e.g. as part of the RSNA haemorrhage detection challenge \cite{rsna2019kaggle} and the CheXpert challenge for automated chest X-Ray interpretation \cite{irvin2019chexpert}. The task of reading the radiology report and assigning labels is not trivial and requires a certain degree of medical knowledge on the part of the human annotator. An alternative is to automatically extract labels, and in this paper we study the task of automatically labelling head computed tomography (CT) radiology reports.

Automatic extraction has traditionally been accomplished using expert medical knowledge to engineer a feature extraction and classification pipeline \cite{yetisgen2013text}; this was the approach taken by Irvin et al. to label the CheXpert dataset of Chest X-Rays \cite{irvin2019chexpert} and by Gorinski et al. in the EdIE-R method for labelling head CT reports \cite{gorinski2019named}. Such pipelines separate the individual tasks such as named entity recognition and negation detection.

An alternative approach is to design an end-to-end machine learning model that will learn to extract the final labels directly from the text. Simple approaches have been demonstrated using word embeddings or bag of words feature representations followed by logistic regression \cite{zech2018natural} or decision trees \cite{yadav-2016-automated}. More complex recurrent neural networks (RNNs) have been shown to be effective for document classification by many authors \cite{yang2016hierarchical,banerjee2019hierarchical} and Drozdov et al. \cite{drozdov2020supervised} show that a bidirectional long short term memory (Bi-LSTM) network with a single attention mechanism also works well for a binary task. However, with recent developments of transformer natural language processing (NLP) models such as Bidirectional Encoder Representations from Transformers (BERT) \cite{devlin2018bert}, it is easier than ever before to use existing pre-trained models that have learnt underlying language patterns and fine-tune them on small domain-specific datasets. This was the approach taken by Wood et al. in the Automated Labelling using an Attention model for Radiology reports of MRI scans (ALARM) model for labelling head magnetic resonance imaging (MRI) reports \cite{wood2020automated}. Specifically, they use BioBERT \cite{alsentzer-etal-2019-publicly} as the base model, which has been pretrained on PubMed abstracts rather than Wikipedia, to obtain contextualised embeddings for each input token and then apply a further attention mechanism to this embedding. Wood et al. perform a binary classification of normal versus abnormal radiology report, which is determined by a number of criteria during data annotation. BERT has also been used for multi-label classification of radiology reports by Smit et al. \cite{smit2020chexbert}. They show that BERT can outperform the previous state of the art for labelling 13 different labels on the CheXpert open source dataset \cite{irvin2019chexpert}. 

Mullenbach et al. proposed per-label attention in a similar document classification task (for clinical coding) in their Convolutional Attention for Multi-Label classification (CAML) model \cite{mullenbach-etal-2018-explainable}. In this paper, inspired by \cite{mullenbach-etal-2018-explainable}, we extend existing state-of-the-art models with a label-dependent attention mechanism. Our contributions are to:
\begin{itemize}
    \item Propose a set of radiographic findings and clinical impressions for labelling of head CT scans for suspected stroke patients.
    \item Show that a multi-headed model with per-label attention improves the accuracy compared to a simple multi-label softmax output.
    \item Show that simple synthetic data significantly improves task performance, especially for classification of rarer labels.
\end{itemize}

\section{Data}
\label{sec:data}

Below we describe the three datasets used in this work.

\subsubsection{NHS GGC dataset:} 
\begin{figure}[h]
    \centering
    \begin{subfigure}{.33\textwidth}
        \centering
        \includegraphics[width=\textwidth]{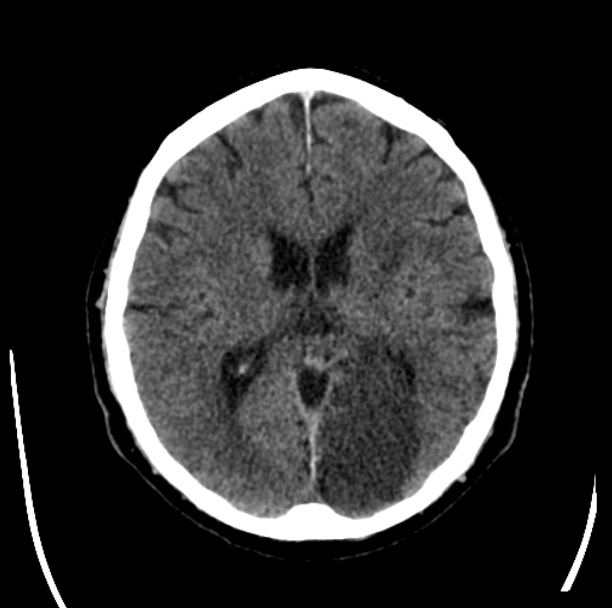}
    \end{subfigure}%
    \begin{subfigure}{.64\textwidth}
        \centering
        \includegraphics[width=\textwidth]{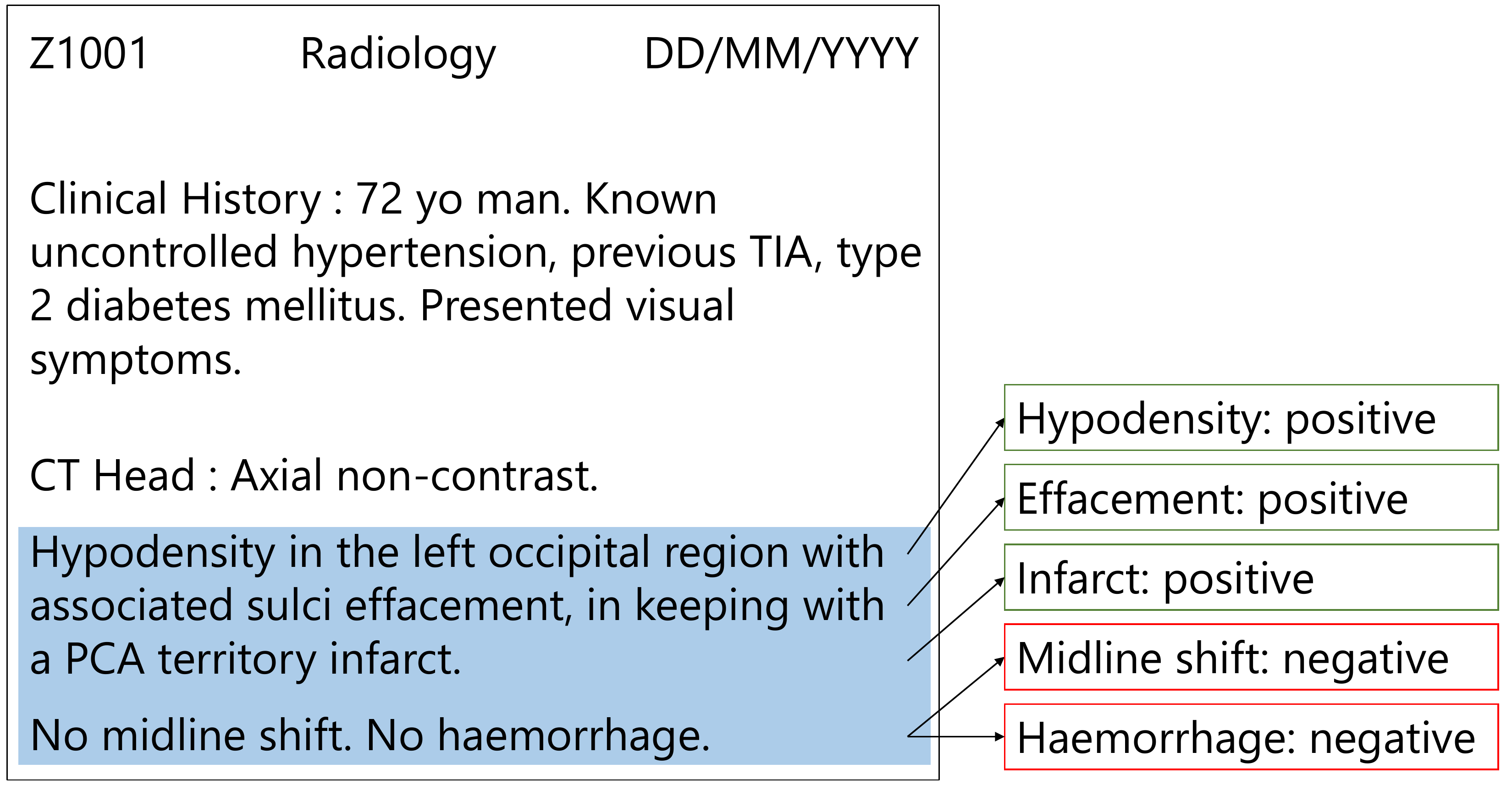}
    \end{subfigure}%
    \caption{Example radiology report. The image (left) shows a slice from an example CT scan\protect\footnotemark; there is a visible darker patch indicating an infarct. The synthetic radiology report (middle) has a similar format to the NHS GGC data. We manually filter relevant sentences, highlighted with blue background. The boxes (right) indicate which labels are annotated.}
    \label{fig:example-report-and-scan}
\end{figure}

Our target dataset contains 230 radiology reports supplied by the NHS Greater Glasgow and Clyde (GGC) Safe Haven. We have the required ethical approval\footnote{iCAIRD project number: 104690; University of St Andrews: CS14871} to use this data. A synthetic example report with similar format to the NHS GGC reports can be seen in Figure~\ref{fig:example-report-and-scan}.

\footnotetext{Case courtesy of Dr David Cuete, Radiopaedia.org, rID: 30225}

\begin{figure}
    \centering
    \includegraphics[width=0.94\textwidth]{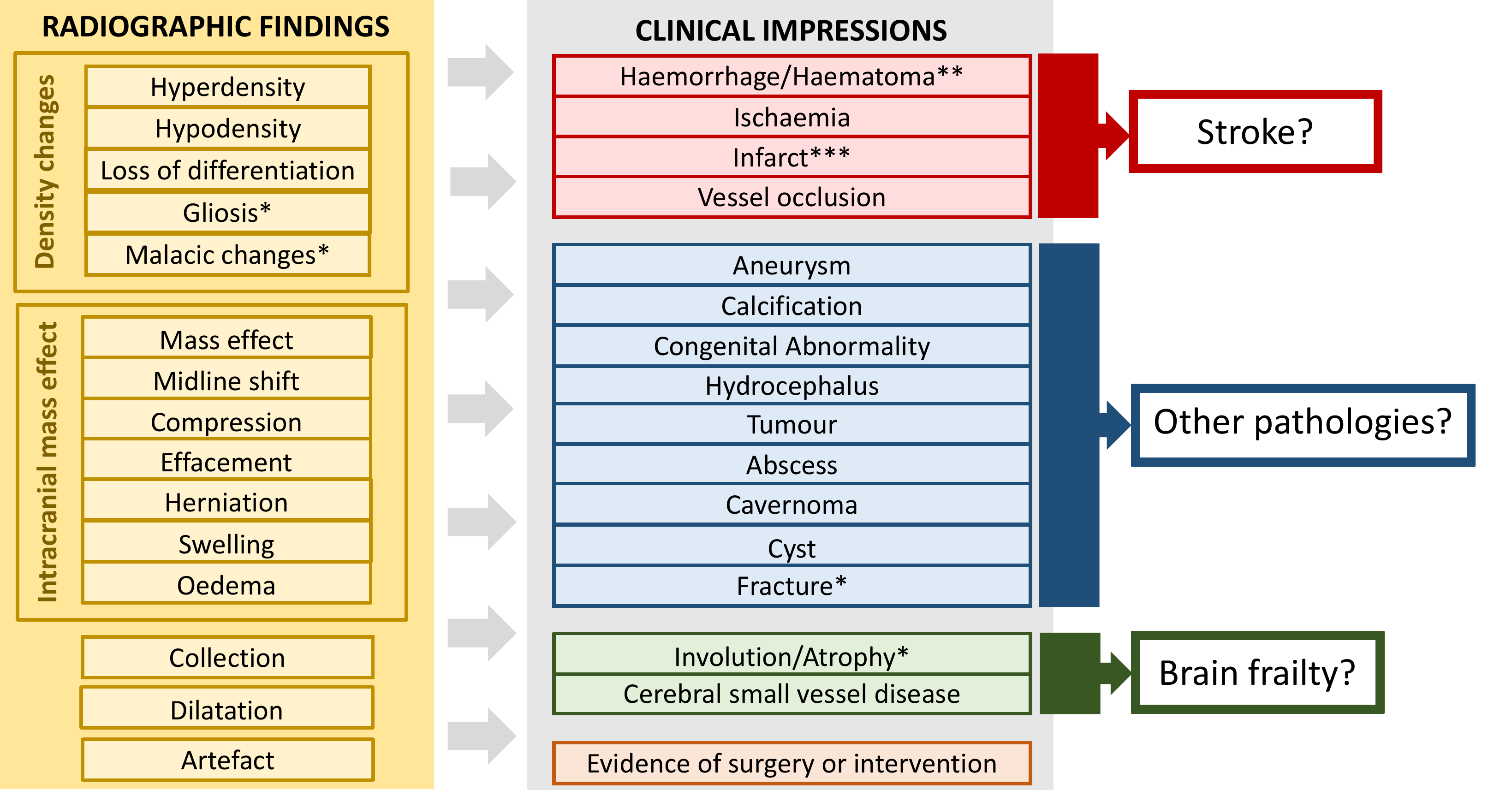}
    \caption{Label schematic: 13 radiographic findings, 14 clinical impressions and 4 crossover labels (finding$\rightarrow$impression links not shown). *These labels fit both the finding and impression categories. **Haematoma can indicate other pathology (e.g. trauma). ***Established infarcts indicate brain frailty \cite{brain-frailty}. }
    \label{fig:labels}
\end{figure}

A list of 31 radiographic findings and clinical impressions found in stroke radiology reports was collated by a clinical researcher; this is the set of labels that we aim to classify. Figure~\ref{fig:labels} shows a complete list of these labels. Each sentence is labelled for each finding or impression as ``positive'', ``uncertain'', ``negative'' or ``not mentioned'' - the same certainty classes as used by Smit et al. \cite{smit2020chexbert}. The most common labels such as ``haemorrhage'', ``infarct'' and ``hyperdensity'' have between 200-400 mentions (100-200 negative, 0-50 uncertain, 100-200 positive) while the rarest labels such as ``abscess'' or ``cyst'' only occur once in the dataset.

During the annotation process, the reports were manually split into sentences by the clinical researcher, resulting in 1,353 sentences which we split into training and validation datasets (due to the limited number of annotated reports, we do not have a separate test set). Each sentence was annotated independently, however we allocate sentences from the same original radiology report to the same dataset to avoid data leakage.

\subsubsection{Synthetic dataset:} We augment our training dataset by synthesising 5 sentences for each label as follows: 
\begin{itemize}
    \item ``There is [label].'' $\rightarrow$ positive
    \item ``There is [label] in the brain.'' $\rightarrow$ positive
    \item ``[Label]'' is evident in the brain.'' $\rightarrow$ positive
    \item ``There may be [label].'' $\rightarrow$ uncertain
    \item ``There is no [label].'' $\rightarrow$ negative
\end{itemize}
For the labels ``haemorrhage/haematoma/contusion'', ``evidence of surgery/ intervention'', ``vessel occlusion (embolus/thrombus)'', and ``involution/atrophy'', we synthesise sentences for each variant. There are 180 synthetic sentences total.

\subsubsection{MIMIC-III dataset:} To pre-train the word embedding, we use clinical notes from the MIMIC dataset \cite{johnson2016mimic}; in total 2,083,180 documents from 46,146 patients.

\noindent{}The datasets are summarised in Table~\ref{data-table}.

\begin{table}[bht]
\centering
\begin{tabular}{lrrr}
\hline
\textbf{Dataset} & \textbf{\#patients} & \textbf{\#reports} & \textbf{\#sentences} \\ 
\hline
NHS GGC -- Training     & 138       & 138       & 838 \\
NHS GGC -- Validation   & 92        & 92        & 515 \\
Synthetic data          & -         & -         & 180 \\
MIMIC-III               & 46,146    & 2,083,180 & 99,718,301 \\
\hline
\end{tabular}
\caption{Summary statistics for the datasets used in this work.}
\label{data-table}
\end{table}

\section{Methods}
\label{sec:methods}

Below we describe the methods which are compared in this paper (implemented in Python). We denote our set of labels as $L$ and our set of certainty classes as $C$, such that the number of labels $n_L = |L|$ and the number of certainty classes $n_C = |C|$. For the NHS GGC dataset, $n_L = 31$ and $n_C = 4$. For all methods, data is pre-processed by extracting sentences and words using the NLTK library \cite{nltk}, removing punctuation, and converting to lower case. Hyperparameter search was performed through manual tuning on the validation set, based on the micro-averaged F1 metric.

\subsection{Simple machine learning approaches}
\label{sec:architectures}

\subsubsection{BoW + RF:}
The Bag of Words + Random Forest (BoW + RF) model uses a bag of words representation as its input. We train one model per label since this gives the most accurate results, resulting in 31 random forest classifiers. Random forest classifiers are quick to train and apply so multiple models are still practical in a real use case. We use the sci-kit learn library \cite{scikit-learn} implementation with 100 estimators, a maximum depth of 10, and 200 maximum features.

\subsubsection{Word2Vec:}
The Word2Vec \cite{mikolov2013distributed} baseline uses a pre-trained word embedding of size $e$. The embedding is pre-trained on the MIMIC dataset described in section~\ref{sec:data} for 30 epochs using the gensim \cite{gensim} library; the vocabulary size is 107,497 words. The word vectors for the input sentence are averaged and passed through a fully connected single layer neural network mapping to an output layer of size $n_L \times n_C$. This network is trained with a constant learning rate of 0.001, batch size of 16 and an embedding size of 200. This and all following models are trained for a maximum of 200 epochs with early stopping patience of 25 epochs on F1 micro. 

\subsection{Deep learning: Per-label attention mechanism}
\label{sec:per-label-attention}

\begin{figure}
    \centering
    \includegraphics[width=\textwidth]{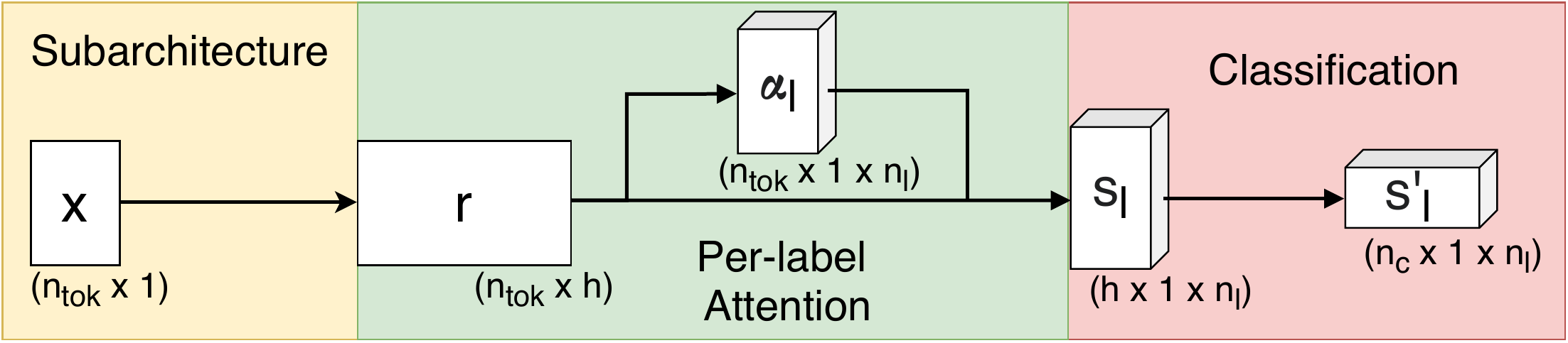}
    \caption{Simplified model diagram: the subarchitecture is a CNN, Bi-GRU or BERT variant and maps from input $x$ to a hidden representation $r$; per-label attention maps to a separate representation $s_l$ for each label before classification.}
    \label{fig:model}
\end{figure}

When training neural networks, we find that accuracy can be reduced where there are many classes. Here we describe the per-label attention mechanism \cite{attention} as seen in Figure~\ref{fig:model}, an adaptation of the multi-label attention mechanism in the CAML model \cite{mullenbach-etal-2018-explainable}. We can apply this to the output of any given neural network subarchitecture. We define the output of the subnetwork as $r \in \mathbb{R}^{n_{tok} \times h}$ where $n_{tok}$ is the number of tokens and $h$ is the hidden representation size. The parameters we learn are the weights $W_0 \in \mathbb{R}^{h \times h}$ and bias $b_0 \in \mathbb{R}^{h}$. Furthermore, for each label $l$ we learn an independent $v_l \in \mathbb{R}^{h}$ to calculate an attention vector $\alpha_l \in \mathbb{R}^{n_{tok}}$.
$$ u = \tanh(W_0 r + b_0) $$
$$ \alpha_l = \text{softmax}(v_l^T u) $$
$$ s_l = \sum \alpha_l r $$
The attended output $s_l \in \mathbb{R}^{h}$ is then passed through $n_L$ parallel classification layers reducing dimensionality from $h$ to $n_C$.

\subsection{Deep learning: Neural network models}

We pre-process the data before input to the neural network architectures. Each input sentence is limited to $n_{tok}$ tokens and padded with zeros to reach this length if the input is shorter. We choose $n_{tok} = 50$ as this is larger than the maximum number of words in any of the sentences in the NHS GGC dataset. The neural network models all finish with $n_L$ softmax classifier outputs, each with $n_C$ classes. 

Models are trained using a weighted categorical cross entropy loss and Adam optimiser \cite{kingma-2015-adam}. We weight across the labels but not across classes, as this did not give any improvements. Given a parameter $\beta$, the number of sentences $n$ and the number of ``not mentioned'' occurrences of a label $o_l$, we calculate the weights for each label using the training data as follows:

$$ w_{l,\text{``not mentioned''}} = \Big ( \frac{n}{o_l} \Big ) ^{\beta} \quad \quad \quad \quad w_{l,\text{``mentioned''}} = \Big ( \frac{n}{n-o_l} \Big ) ^{\beta} $$

\subsubsection{CAML:}
The CAML model follows the implementation by Mullenbach et al. \cite{mullenbach-etal-2018-explainable} and uses an embedding that is initialised to the same pre-trained weights as for the Word2Vec baseline. The embedded input passes through a convolutional layer of graduated filter sizes applied in parallel (see below), followed by max-pooling operations across each graduated set of filters, to produce our intermediate representation $r$. This is then passed through the per-label attention mechanism introduced by the CAML model. For the convolutional layer, we chose 512 CNN filter maps with kernel sizes of 2 and 4. The model was trained with a learning rate of 0.0005 and a batch size of 16.

\subsubsection{Bi-GRU:}
The embedding is initialised to the same pre-trained weights as used for the Word2Vec baseline. The embedded sentence $x$ passes through a bidirectional GRU (Bi-GRU) network \cite{cho-etal-2014-properties} with hidden size of $h/2$. The outputs from both directions are concatenated to produce a representation $r$ for each input sentence. For \textbf{Bi-GRU + single attention}, this representation is passed through a single attention mechanism. For \textbf{Bi-GRU + per-label attention}, this representation is passed through the per-label attention mechanism. The model was trained with a learning rate of 0.0005, batch size of 16 and hidden size $h = 1024$.

\subsubsection{BERT and BioBERT:}
The BERT model is a standard pre-trained BERT model, ``bert-base-uncased'' - weights are available for download online\footnote{ \url{https://github.com/google-research/bert}} - we use the huggingface \cite{huggingface} implementation.
We take the output representation for the CLS token of size $768\times1$ at position 0 and follow with the $n_L$ softmax outputs. The model was trained with a learning rate of 0.0001 and batch size of 32. For \textbf{BioBERT}, we use a Bio-/ClinicalBERT model pretrained on both PubMed abstracts and the MIMIC-III dataset\footnote{ \url{https://github.com/th0mi/clinicalBERT} } with the huggingface BERT implementation. We use the same training parameters as for BERT (above). 

\subsubsection{ALARM:}
Our implementation of the ALARM \cite{wood2020automated} model uses the BioBERT model (and training parameters) described above. Following the implementation details of Wood et al., instead of using a single output vector of size $768\times1$, we extract the entire learnt representation of size $768 \times n_{tok}$. For the \textbf{ALARM + softmax} model, we pass this through a single attention vector and then through three fully connected layers to map from $768$ to $512$ to $256$ to the $n_L \times n_C$ outputs. For \textbf{ALARM + per-label-attention}, we employ $n_L$ per-label attention mechanisms instead of a single shared attention mechanism before passing through three fully connected layers \textit{per label}.

\section{Results}

Tables \ref{tab:micro-F1} and \ref{tab:macro-F1} show the results. We report the micro-averaged F1 score as our main metric, calculated across all labels. We also report the macro-averaged F1 score; this is F1 score averaged across all labels with equal weighting for each label.
We note that although we used micro F1 as our early stopping criterion, we do not observe an obvious difference in the scores if F1 macro is used for early stopping. We exclude the ``not mentioned'' certainty from our metrics, similar to the approach used by Smit et al. \cite{smit2020chexbert} - we denote $C' = C\backslash\{``not\ mentioned"\}$, so $n_{c'} = n_C-1$. When we report our F1 metrics for a single certainty class we report the usual F1 metric, whereas when we report metrics for all classes and labels we report an average per certainty class.

For all experiments, we use a machine with NVIDIA GeForce GTX 1080 Ti GPU (11GB of VRAM), Intel Xeon CPU E5 v3 (6 physical cores, maximum clock frequency of 3.401 GHz) and 32GB of RAM. Training run times range from 14 seconds for the Random Forest model to 376 seconds for the Bi-GRU + per-label attention model and 1448 seconds for the ALARM + per-label-attention model. For details of all run times, see Table~\ref{timings-table}.

\begin{table}[!h]
\centering
\begin{tabular}{lcccc}
\hline
\textbf{Model} & \textbf{All} & \textbf{Negative} & \textbf{Uncertain} & \textbf{Positive} \\ 
\hline
BoW + RF & $0.871_{0.003}$ & $0.936_{0.003}$ & $0.119_{0.021}$ & $0.889_{0.003}$ \\
Word2Vec & $0.808_{0.005}$ & $0.900_{0.007}$ & $0.328_{0.023}$ & $0.812_{0.008}$ \\
CAML \cite{mullenbach-etal-2018-explainable} & $0.838_{0.005}$ & $0.866_{0.011}$ & $0.135_{0.050}$ & $0.873_{0.001}$ \\
Bi-GRU & $0.868_{0.009}$ & $0.936_{0.011}$ & $0.488_{0.0.017}$ & $0.872_{0.009}$ \\
Bi-GRU + single attention & $0.863_{0.009}$ & $0.924_{0.017}$ & $0.424_{0.032}$ & $0.873_{0.006}$ \\
Bi-GRU + per-label attention & $0.921_{0.003}$ & $\bm{0.970}_{0.006}$ & $0.573_{0.011}$ & $0.932_{0.004}$ \\
BERT & $0.907_{0.003}$ & $0.953_{0.004}$ & $0.585_{0.035}$ & $0.916_{0.002}$ \\
BioBERT & $0.915_{0.005}$ & $0.959_{0.003}$ & $0.627_{0.040}$ & $0.922_{0.007}$ \\
ALARM + softmax & $0.899_{0.008}$ & $0.948_{0.002}$ & $0.570_{0.028}$ & $0.909_{0.010}$ \\
ALARM + per-label attention & $\bm{0.928}_{0.008}$ & $0.965_{0.004}$ & $\bm{0.689}_{0.039}$ & $\bm{0.936}_{0.008}$ \\
\hline
\end{tabular}
\caption{\label{micro-f1-results-table} Micro-averaged F1 results as $\text{mean}_{\text{standard deviation}}$ of 5 runs with different random seeds. ``All'' combines the classes ``negative'', ``uncertain'' and ``positive''. Bold indicates the best model for each metric.}
\label{tab:micro-F1}
\end{table}

\FloatBarrier

\begin{table}[!ht]
\centering
\begin{tabular}{lcccc}
\hline
\textbf{Model} & \textbf{All} & \textbf{Negative} & \textbf{Uncertain} & \textbf{Positive} \\ 
\hline
BoW + RF & $0.477_{0.013}$ & $0.667_{0.019}$ & $0.052_{0.025}$ & $0.711_{0.001}$ \\
Word2Vec & $0.455_{0.011}$ & $0.581_{0.034}$ & $0.164_{0.048}$ & $0.619_{0.029}$ \\
CAML \cite{mullenbach-etal-2018-explainable} & $0.394_{0.013}$ & $0.435_{0.017}$ & $0.086_{0.050}$ & $0.661_{0.025}$ \\
Bi-GRU & $0.631_{0.025}$ & $0.718_{0.042}$ & $0.404_{0.051}$ & $0.718_{0.011}$ \\
Bi-GRU + single attention & $0.522_{0.039}$ & $0.666_{0.065}$ & $0.223_{0.051}$ & $0.677_{0.018}$ \\
Bi-GRU + per-label attention & $0.708_{0.014}$ & $0.796_{0.027}$ & $0.524_{0.023}$ & $0.803_{0.016}$ \\
BERT & $0.673_{0.015}$ & $0.773_{0.004}$ & $0.457_{0.050}$ & $0.790_{0.025}$ \\
BioBERT & $0.673_{0.041}$ & $0.730_{0.038}$ & $0.529_{0.094}$ & $0.761_{0.017}$ \\
ALARM + softmax & $0.652_{0.025}$ & $0.767_{0.009}$ & $0.441_{0.071}$ & $0.749_{0.007}$ \\
ALARM + per-label attention & $\bm{0.766}_{0.028}$ & $\bm{0.818}_{0.029}$ & $\bm{0.661}_{0.061}$ & $\bm{0.818}_{0.021}$ \\
\hline
\end{tabular}
\caption{\label{macro-f1-results-table} Macro-averaged F1 results as $\text{mean}_{\text{standard deviation}}$ of 5 runs with different random seeds. ``All'' combines the classes ``negative'', ``uncertain'' and ``positive''. Bold indicates the best model for each metric.}
\label{tab:macro-F1}
\end{table}

\FloatBarrier
\subsubsection{Per-label attention:}
The micro- and macro-averaged F1 scores (Tables~\ref{micro-f1-results-table} and \ref{macro-f1-results-table}) show that for both BioBERT and the Bi-GRU models, adding \emph{per-label} attention to the models improves performance consistently over the models with a single attention mechanism (p-values of $<0.05$). We also show the breakdown in accuracies across certainty classes (negative, uncertain and positive) in our results tables. It can be seen that the per-label attention provides large gains in accuracy across all classes. The macro F1 metric amplifies this because all labels are weighted equally, giving an idea of how the model performs for the rarer labels, several of which have fewer than 10 training samples each. 

\begin{figure}[th]
    \centering
    \includegraphics[width=\textwidth]{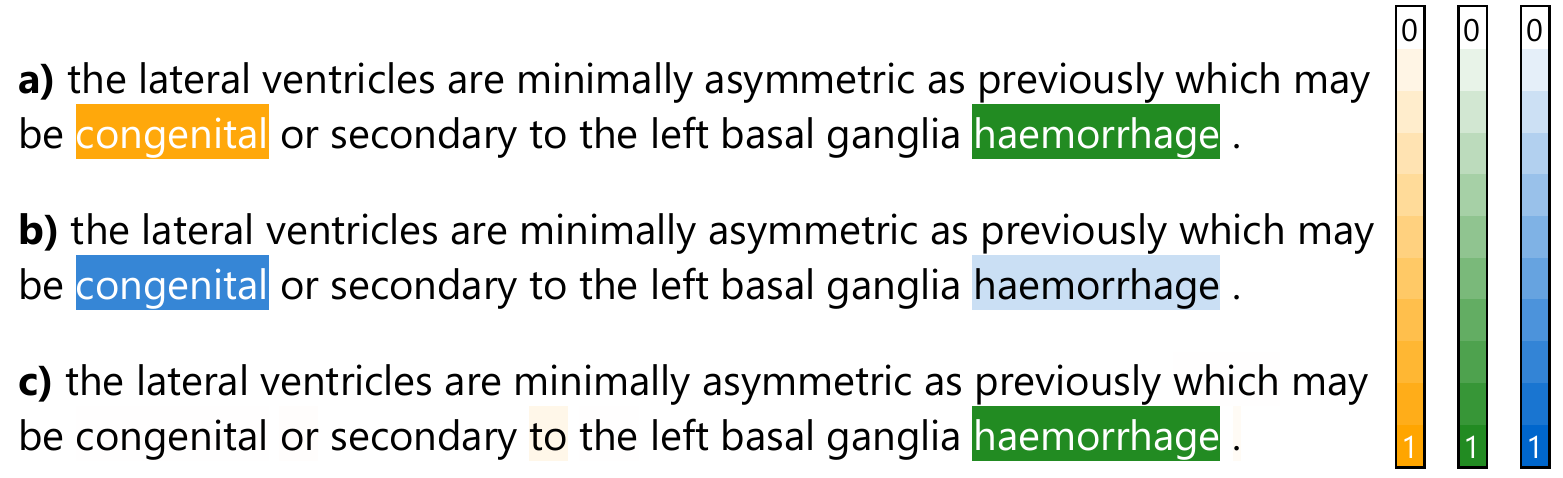}
    \caption{Visualisation of attention for (a) per-label attention vectors, (b) a single attention vector and (c) per-label attention from a model trained without synthetic data. Model (a) detects congenital (yellow) and haemorrhage (green) separately. Model (b) detects both keywords in the single attention vector (blue). Model (c) does not detect the ``congenital'' keyword. }
    \label{fig:attention}
\end{figure}

Figure~\ref{fig:attention} compares the attention learnt by a single attention model to per-label attention models. We see that the single attention vector (Figure~\ref{fig:attention}b) attends to the correct words - ``congenital'' and ``haemorrhage'' - however the model incorrectly predicts both labels as ``not mentioned''. In comparison, the model with per-label attention (Figure~\ref{fig:attention}a) recognises the same keywords separately within the respective label attention mechanisms, and correctly predicts both labels as ``positive''. This makes sense because the single attention mechanism does not have separate follow-on $s_l$ representations and therefore features for all labels are entangled in one representation. Finally, the model trained without synthetic data (Figure~\ref{fig:attention}c) does not recognise the ``congenital'' keyword and does not make the correct prediction for this label. 

\subsubsection{Synthetic data and importance of pre-training:}
\begin{figure}[t]
    \centering
    \includegraphics[width=\textwidth]{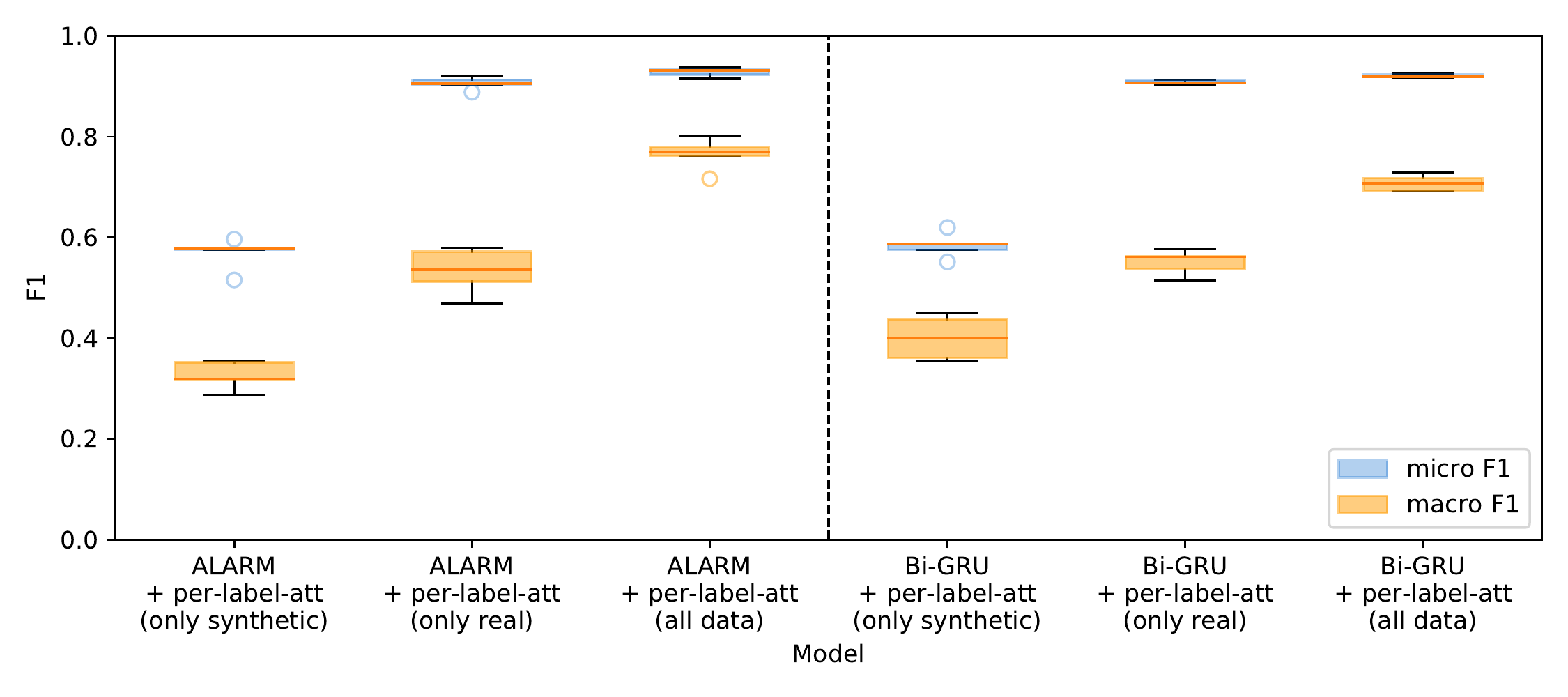}
    \caption{Graph showing effect of synthetic data on micro-averaged F1 (blue) and macro-averaged F1 (orange). Synthetic data gives consistent improvement.}
    \label{fig:synthetic-data}
\end{figure}

To investigate the effect of the synthetic training data, we train models on only the synthetic data, only NHS GGC data, and both combined. The results for macro F1 in Figure~\ref{fig:synthetic-data} clearly show an improvement when the synthetic data is used alongside the original data - this is consistent across both of our best models (p-values of $<0.05$). For numerical results see Tables~\ref{tab:ablations-micro} and \ref{tab:ablations-macro}.

We also investigated the effect of the embedding pre-training. A model with randomly initialised embeddings (maintaining the same vocabulary and embedding size) performs 0.028 worse for the micro-averaged F1 compared to a model using a pre-trained embedding (p-value of $<0.05$).

\subsubsection{Error Analysis:}
When investigating the prediction errors of our best model, we identify that approximately 30\% of errors are due to missed labels, 10\% are due to falsely predicted labels, and the remaining 60\% are due to confusion between certainty classes (negative, uncertain, positive). Many of the missed labels are caused by previously unseen synonyms or subtypes, for instance ``arteriovenous malformation'' is an instance of ``congenital abnormality'' which is a diverse class. There are also many ways of expressing certainty which are subtly different; for instance positive might be expressed as ``probable'', ``likely'', ''indicates'', ''suggestive of'', ''is consistent with'' whereas uncertainty might be expressed as ``possible'', ''may represent'', ''could indicate'', ''is suspicious of'' and other subtly different expressions. Errors might be mitigated with the use of a larger training dataset and richer data synthesis, potentially by exploiting medical knowledge bases such as UMLS \cite{umls} to augment the synthetic dataset with a rich synonym set.

\section{Conclusions and Future Work}

We have introduced a set of radiographic findings and clinical impressions that are relevant for stroke and can be extracted from head CT radiology reports. For deep learning approaches, we have shown that per-label attention and a simple synthetic dataset each improve accuracy for our multi-label classification task, yielding a recipe for scalable learning of many labels. In future work, we intend to annotate a larger dataset as well as  leveraging knowledge bases to create a richer synthetic dataset. Furthermore, the labels generated by our models should be used to train an image analysis algorithm on the associated head CT scans.

\section{Acknowledgements}

This work is part of the Industrial Centre for AI Research in digital Diagnostics (iCAIRD) which is funded by Innovate UK on behalf of UK Research and Innovation (UKRI) [project number: 104690].
We would like to thank the Glasgow Safe Haven for assistance in creating and providing this dataset.
Thanks also to The Data Lab for support and funding.

\clearpage

\appendix
\section*{Appendix}

\begin{table}
\centering
\begin{tabular}{lccc}
\hline
 & & \textbf{Training} & \textbf{Inference time} \\ 
\textbf{Model} & \textbf{\#Parameters} & \textbf{time [s]} & \textbf{[s/sample]} \\ 
\hline
BoW + RF & n/a & $14_1$ & $0.2933_{0.0040}$ \\
Word2Vec & $166,524$ & $46_{8}$ & $0.0022_{0.0001}$ \\
CAML \cite{mullenbach-etal-2018-explainable} & $1,021,176$ & $250_{43}$ & $0.0099_{0.0008}$ \\
Bi-GRU & $2,889,852$ & $111_{32}$ & $0.0066_{0.0003}$ \\
Bi-GRU + single attention & $3,371,132$ & $120_{55}$ & $0.0062_{0.0003}$ \\
Bi-GRU + per-label attention & $3,401,852$ & $376_{127}$ & $0.0109_{0.0004}$ \\
BERT & $109,577,596$ & $1115_{322}$ & $0.0565_{0.0025}$ \\
BioBERT & $109,577,596$ &  $927_{171}$ & $0.0575_{0.0008}$ \\
ALARM + softmax & $109,458,556$ & $911_{243}$ & $0.0590_{0.0013}$ \\
ALARM + per-label attention & $125,233,276$ & $1448_{375}$ & $0.0740_{0.0002}$ \\
\hline
\end{tabular}
\vspace{0.1cm}
\caption{\label{timings-table} Number of parameters, training time (over 838 samples) and inference time (per sample) for all models. All timings are given as $\text{mean}_{\text{standard deviation}}$ of 5 runs with different random seeds. The fastest model to train is the random forest model. The Bi-GRU network is significantly faster to train than BERT \cite{devlin2018bert} and ALARM \cite{wood2020automated} due to the smaller number of parameters. The only model that is faster than the Bi-GRU model is Word2Vec which has a far inferior F1 score. The random forest model is the slowest at inference time because it has $n_L$ models (one model per label) - the inference could be parallelised to improve performance.}
\end{table}

\begin{table}
\centering
\begin{tabular}{lcccccc}
\hline
\textbf{Model} &
\textbf{Embedding} & \textbf{Data} & \textbf{All} & \textbf{Negative} & \textbf{Uncertain} & \textbf{Positive} \\ 
\hline
Bi-GRU & MIMIC & S & $0.584_{0.022}$ & $0.496_{0.089}$ & $0.204_{0.031}$ & $0.642_{0.012}$ \\
Bi-GRU & MIMIC & N-S & $0.908_{0.004}$ & $0.956_{0.004}$ & $0.427_{0.058}$ & $0.927_{0.004}$ \\
Bi-GRU & Random & N+S & $0.893_{0.002}$ & $0.962_{0.008}$ & $0.432_{0.033}$ & $0.903_{0.002}$ \\
Bi-GRU & MIMIC & N+S & $0.921_{0.003}$ & $\bm{0.970}_{0.006}$ & $0.573_{0.011}$ & $0.932_{0.004}$ \\
ALARM & MIMIC & S & $0.569_{0.028}$ & $0.725_{0.062}$ & $0.128_{0.041}$ & $0.531_{0.028}$ \\
ALARM & MIMIC & N-S & $0.906_{0.011}$ & $0.944_{0.005}$ & $0.532_{0.087}$ & $0.923_{0.010}$ \\
ALARM & MIMIC & N+S & $\bm{0.928}_{0.008}$ & $0.965_{0.004}$ & $\bm{0.689}_{0.039}$ & $\bm{0.936}_{0.008}$ \\
\hline
\end{tabular}
\vspace{0.1cm}
\caption{\label{tab:ablations-micro} Results for our ablation studies showing \textit{micro-averaged} F1 as $\text{mean}_{\text{standard deviation}}$ of 5 runs with different random seeds (all models are trained with per-label attention). \textit{N} data is the NHS GGC dataset and \textit{S} is the synthetic dataset. ``All'' combines the classes ``negative'', ``uncertain'' and ``positive''. Bold indicates the best model for each metric.}
\end{table}

\begin{table}
\centering
\begin{tabular}{lcccccc}
\hline
\textbf{Model} &
\textbf{Embedding} & \textbf{Data} & \textbf{All} & \textbf{Negative} & \textbf{Uncertain} & \textbf{Positive} \\ 
\hline
Bi-GRU & MIMIC & S & $0.400_{0.039}$ & $0.504_{0.050}$ & $0.106_{0.029}$ & $0.590_{0.065}$ \\
Bi-GRU & MIMIC & N-S & $0.551_{0.024}$ & $0.623_{0.024}$ & $0.268_{0.089}$ & $0.761_{0.026}$ \\
Bi-GRU & Random & N+S & $0.617_{0.015}$ & $0.746_{0.042}$ & $0.360_{0.054}$ & $0.745_{0.024}$ \\
Bi-GRU & MIMIC & N+S & $0.708_{0.014}$ & $0.796_{0.027}$ & $0.524_{0.023}$ & $0.803_{0.016}$ \\
ALARM & MIMIC & S & $0.326_{0.025}$ & $0.607_{0.039}$ & $0.065_{0.032}$ & $0.307_{0.021}$ \\
ALARM & MIMIC & N-S & $0.534_{0.041}$ & $0.598_{0.027}$ & $0.245_{0.088}$ & $0.758_{0.038}$ \\
ALARM & MIMIC & N+S & $\bm{0.766}_{0.028}$ & $\bm{0.818}_{0.029}$ & $\bm{0.661}_{0.061}$ & $\bm{0.818}_{0.021}$ \\
\hline
\end{tabular}
\vspace{0.1cm}
\caption{\label{tab:ablations-macro}  Results for our ablation studies showing \textit{macro-averaged} F1 as $\text{mean}_{\text{standard deviation}}$ of 5 runs with different random seeds (all models are trained with per-label attention). \textit{N} data is the NHS GGC dataset and \textit{S} is the synthetic dataset. ``All'' combines the classes ``negative'', ``uncertain'' and ``positive''. Bold indicates the best model for each metric.}
\end{table}

\bibliographystyle{splncs04}
\bibliography{paper11}

\end{document}